%% file: iclr2025_conference.tex
\documentclass{article}
\usepackage[T1]{fontenc}
\usepackage{iclr2027_conference,times}
\input{math_commands.tex}

\usepackage{graphicx}
\usepackage{hyperref}
\usepackage{url}
\hypersetup{hidelinks}
\usepackage{placeins}
\usepackage{booktabs}
\usepackage{tabularx}
\usepackage{array}
\usepackage{multirow}
\definecolor{tableink}{HTML}{242A30}
\definecolor{tableaccent}{HTML}{486B80}
\definecolor{tablepaper}{HTML}{F6F8F9}
\title{EvalResearchBench: Can AI Agents Build Their Own Feedback for Recursive Self-Improvement?}
\author{%
Yaolun Zhang\textsuperscript{1,4,*}\quad Tianyi Xu\textsuperscript{1,*}\quad Yujie Zhao\textsuperscript{2,*}\\
\bfseries Jishen Zhao\textsuperscript{2}\quad Qingyun Wu\textsuperscript{3,4}\quad Huazheng Wang\textsuperscript{1,4}\\[0.35em]
\normalfont\small\textsuperscript{1}Oregon State University\quad \textsuperscript{2}University of California, San Diego\\
\normalfont\small\textsuperscript{3}Pennsylvania State University\quad \textsuperscript{4}AG2 AI\\[0.2em]
\normalfont\small\textsuperscript{*}Equal contribution.
}
\iclrfinalcopy
\begin{document}
\raggedbottom
\maketitle
\lhead{Preprint}
\hypersetup{pdftitle={EvalResearchBench: Can AI Agents Build Their Own Feedback for Recursive Self-Improvement?},pdfauthor={Yaolun Zhang, Tianyi Xu, Yujie Zhao, Jishen Zhao, Qingyun Wu, Huazheng Wang}}
\input{sections/abstract}
\vspace{-0.5em}
\begin{center}
\small
\href{https://mercury7353.github.io/self-evaluation-bench/}{\raisebox{-0.18em}{\includegraphics[height=1em]{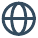}}\enspace Project Website}
\hspace{2em}
\href{https://github.com/Mercury7353/self-evaluation-bench}{\raisebox{-0.18em}{\includegraphics[height=1em]{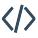}}\enspace GitHub Code}
\end{center}
\vspace{-0.75em}
\input{figures/figure1}
\clearpage
\input{sections/introduction}
\input{sections/related_work}
\input{sections/method}
\input{sections/experiments}
\input{sections/agent_behavior_analysis}
\input{sections/conclusion}
\clearpage
\bibliography{references}
\bibliographystyle{iclr2027_conference}
\clearpage
\appendix
\input{sections/appendix_comparison}

\input{sections/appendix_ablations}
\FloatBarrier
\input{sections/appendix_research_components}
\input{sections/appendix_prompts}
\FloatBarrier
\input{sections/appendix_research_case}
\end{document}

%% file: math_commands.tex
\usepackage{amsmath,amsfonts,bm}

\def\eqref#1{equation~\ref{#1}}

\def\1{\bm{1}}

\DeclareMathAlphabet{\mathsfit}{\encodingdefault}{\sfdefault}{m}{sl}
\SetMathAlphabet{\mathsfit}{bold}{\encodingdefault}{\sfdefault}{bx}{n}



%% file: sections/abstract.tex
\begin{abstract}
Recursive self-improvement (RSI) relies on evaluation feedback to assess progress and guide further research, yet repeatedly running complex benchmarks is costly and slows iteration. Human experts reduce this cost by selecting benchmark subsets or designing compact suites. We ask whether AI agents can automate this design process and introduce EvalResearchBench (ERB), a benchmark for autonomous evaluation research. Given target materials, development references, candidate APIs, and fixed time and API budgets, an agent called the researcher selects or synthesizes tasks, implements graders, and revises them in pilot tests before freezing an executable evaluator for coding, co-work, and reasoning. We study 9 researchers and 13 candidate models and compare each frozen evaluator with 14 target benchmarks on score concordance and pairwise agreement. The best evaluators order about 75\% of candidate pairs as the targets do, below the 91\% ceiling set by disagreements among the targets. No researcher leads on every metric, and the best evaluator on development targets is not the best on sealed targets hidden from the researcher. A human-designed sample of public tasks remains a strong baseline, and the evaluator with the lowest recorded execution cost attains the highest pairwise agreement. Agents repair tasks and graders through pilot feedback, yet their evaluators can still truncate answers, exhaust the evaluation budget, or let a few questions dominate a domain score.
\end{abstract}

%% file: figures/figure1.tex
\begin{figure}[!h]
\centering
\includegraphics[width=\linewidth]{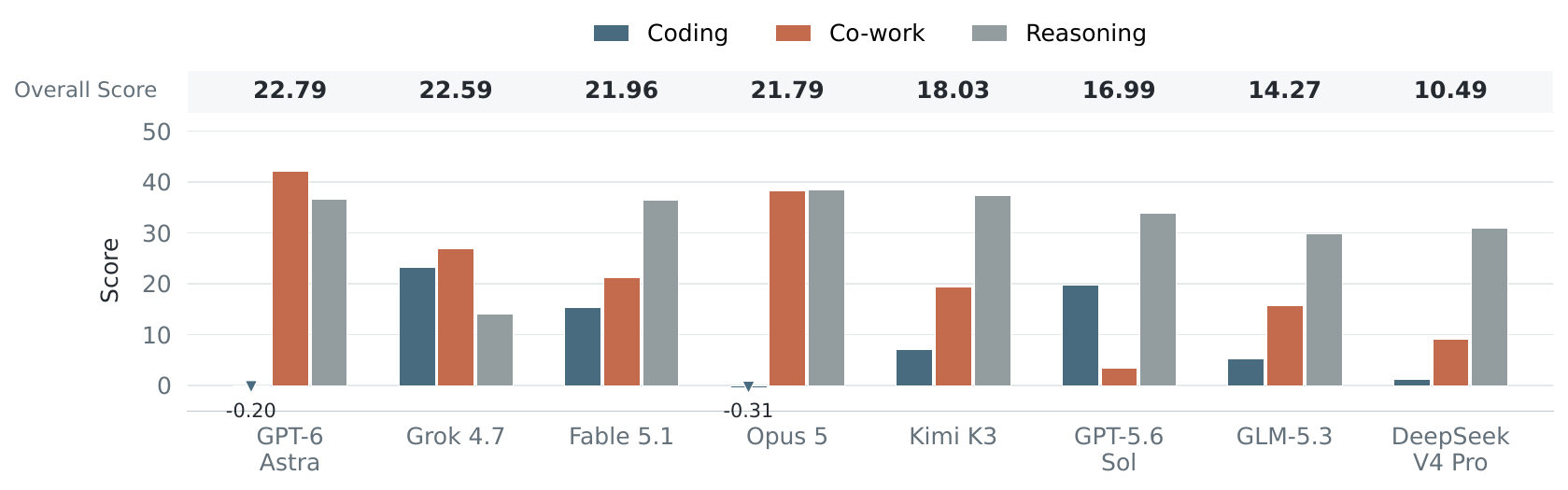}
\vspace{-8pt}
\caption{\textbf{Evaluation designs have distinct domain strengths.} Eight assessed programs, ordered by overall Score. Grouped bars share the same scale; overall averages all fourteen targets equally.}
\vspace{-10pt}
\label{fig:researcher-domains}
\end{figure}

%% file: sections/introduction.tex
\section{Introduction}
\label{sec:introduction}

Recursive self-improvement (RSI) relies on evaluation feedback to determine whether a change improves a system and which direction to explore next. Each new training recipe, dataset, or program creates another version to assess. Repeatedly running complex benchmarks incurs API and compute costs and delays the next experiment, particularly when evaluation requires long tool-using trajectories or specialized environments. Human experts reduce this burden by selecting representative subsets or designing cheaper proxies, as in tinyBenchmarks~\citep{polo2024tinybenchmarks}. An autonomous improvement loop would need to perform this design step itself and obtain feedback that is informative enough to guide decisions and cheap enough to repeat.

Recent work automates improvement while relying on supplied evaluation protocols. In PostTrainBench, agents choose data and post-training algorithms, but the target benchmark and evaluation harness are fixed~\citep{rank2026posttrainbench}. RSIBench-Data lets agents revise training-data strategies using checkpoint scores and trajectories from a fixed evaluation service~\citep{meng2026rsibenchdata}. AlphaEvolve evolves programs using a user-provided evaluation function, with optional user-specified test cascades to screen proposals efficiently~\citep{novikov2025alphaevolve}. In each case, humans decide what to test, how to score it, and how to keep repeated measurement affordable, so extending autonomy to a new domain still requires a human-designed evaluation.

Other work automates parts of evaluation design. PACE selects inexpensive source questions and calibrates them against expensive agentic targets~\citep{song2026pace}. AutoBencher and BenchAgents automate benchmark construction and refinement~\citep{li2024autobencher,butt2024benchagents}, and BenchBench compares the quality of model-designed suites~\citep{zheng2026benchbench}. None of these settings poses the full research problem, in which an agent chooses a measurement strategy, spends a limited experimental budget, and revises an executable evaluator until it tracks complex target benchmarks. Solving it requires inspecting target evidence, choosing or creating tasks, verifying graders, and deciding whether a pilot failure calls for harder questions, different coverage, a repaired rubric, or more execution budget. We therefore ask whether an agent with fixed time and API budgets can build an evaluator that reproduces how complex target benchmarks score and rank models. We call this capability autonomous evaluation research.

To address these gaps, we introduce \textbf{EvalResearchBench (ERB)}, in which the researcher designs the measurement itself, pays for every pilot from a fixed budget, and submits an executable evaluator that is scored against complex target benchmarks. A \emph{researcher} is a model in an agent harness that designs the evaluator, a \emph{candidate} is a model being tested, and a \emph{target benchmark} is an established evaluation whose candidate scores the evaluator should reproduce. The researcher receives target descriptions, permitted tasks and grading resources, execution traces, development reference scores, and APIs for development candidates (Figure~\ref{fig:pipeline}). In pilot tests, it observes candidate responses, costs, and agreement with the visible references, then revises its tasks, graders, weights, and answer allowances. It then freezes an evaluator that returns one score per domain. Sealed candidates and targets stay hidden during research, so the researcher can neither call those candidates nor see those reference scores. Independent assessment runs the frozen evaluator on all candidates under a separate budget and compares its scores with both development and sealed targets.

\input{figures/figure2}

We evaluate 9 researcher configurations, each a model in the Codex or Claude Code harness run for one episode, on 13 open-weight and proprietary candidates and 14 target benchmarks in coding, co-work, and reasoning (Table~\ref{tab:setup}). These domains cover the implementation, professional work, and analysis that autonomous research relies on. As a baseline, Random120 samples public tasks under fixed source quotas designed by humans.

We measure agreement in two ways. \emph{Score} is 100 times Lin's concordance correlation coefficient between measured and reference scores~\citep{lin1989concordance}, so it rewards matching score level and spread as well as order. \emph{Pairwise agreement} (PA) is the fraction of candidate pairs that the evaluator orders as the target does. We also separate the one-time cost of building an evaluator from the cost of running it, which recurs for every new candidate.

No evaluator leads in every domain (Figure~\ref{fig:researcher-domains}). Grok~4.7 leads coding, Astra leads co-work, and Opus~5 leads reasoning, while coding is the weakest domain for most evaluators. Astra attains the highest overall Score and Grok~4.7 the highest PA. Even the best PA, about 75\%, stays well below the 91\% ceiling set by disagreements among the targets. Fable~5.1 has the highest Score on development targets but not on sealed ones, so fitting visible references does not ensure transfer. Random120 remains a strong baseline. On identical candidates, seven of eight agent-designed evaluators exceed it in Score, but only four exceed it in PA. Evaluators with similar construction cost differ markedly in execution cost, and the one with the lowest recorded execution cost attains the highest PA. Agents repair tasks and graders through pilot feedback, yet their final evaluators can still truncate answers, exhaust the assessment budget, or let a few heavily weighted questions dominate a domain score.

%% file: figures/figure2.tex
\begin{figure}[!t]
\centering
\includegraphics[width=\linewidth]{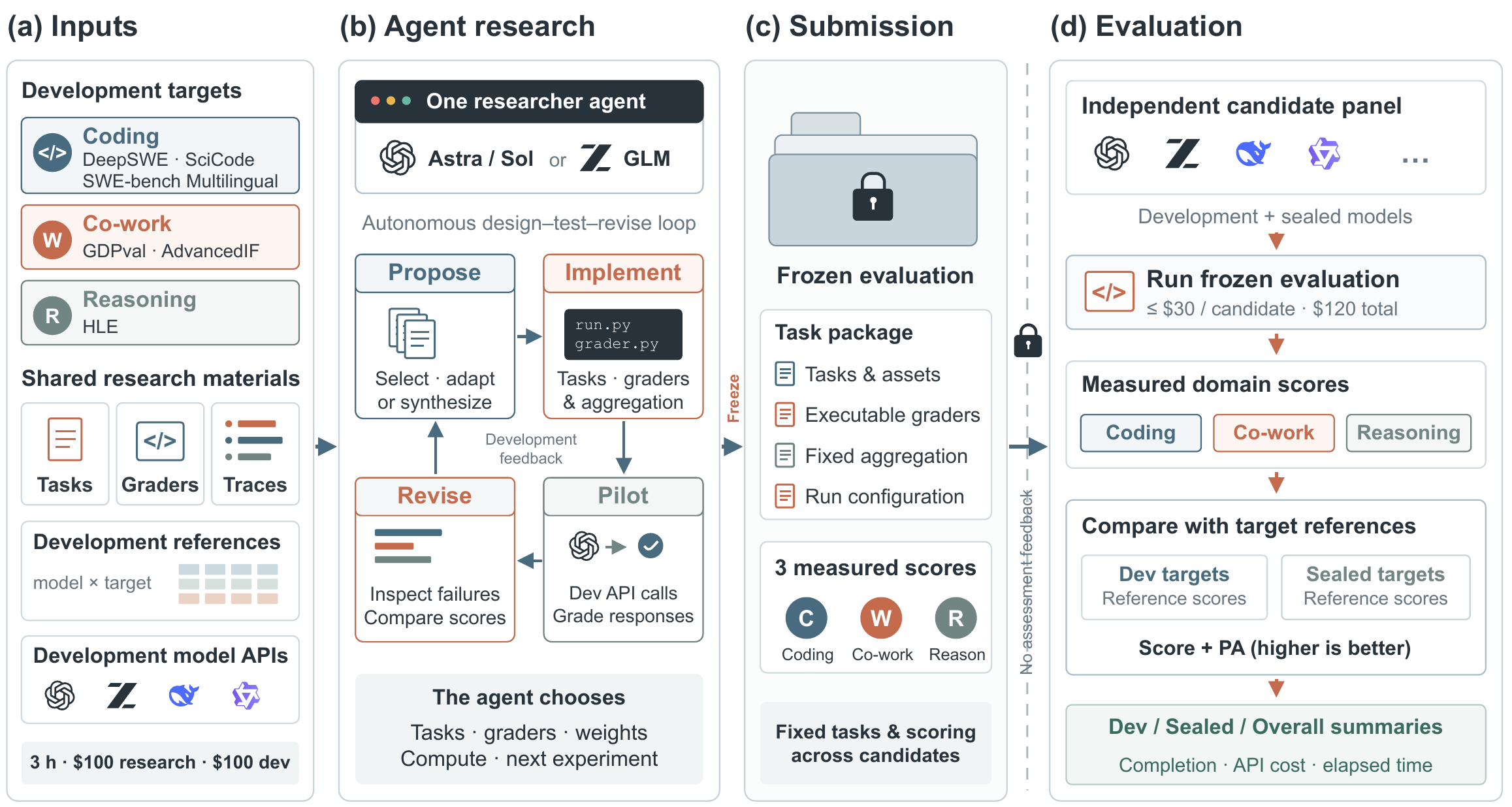}
\caption{\textbf{EvalResearchBench.} Research materials support an autonomous design--test--revise loop. Tasks, graders, and score weights are frozen before independent assessment. Candidate-model and target visibility are assigned separately.}
\label{fig:pipeline}
\end{figure}

%% file: sections/related_work.tex
\section{Related Work}
\label{sec:related-work}

\subsection{Evaluation Design and Efficient Measurement}
\label{sec:related-evaluation}
Evaluation design spans coverage and standardization~\citep{ribeiro2020checklist,liang2022helm,kiela2021dynabench}, task construction~\citep{perez2022modelwritten,liu2023evalplus,xia2024evoeval}, and automated grading~\citep{liu2023geval,zheng2023judge}. AutoBencher refines datasets using model feedback~\citep{li2024autobencher}; BenchAgents integrates planning, generation, verification, and evaluation~\citep{butt2024benchagents}; BenchBench assesses benchmark designers through validity, discrimination, and ranking reliability~\citep{zheng2026benchbench}. Efficient measurement offers another route: APTBench transforms agent tasks and trajectories into lightweight questions~\citep{qin2025aptbench}, while tinyBenchmarks, Fluid Benchmarking, and Efficient Benchmarking of AI Agents study selection, estimation, and adaptive testing~\citep{polo2024tinybenchmarks,hofmann2025fluid,ndzomga2026efficient}. Agent Psychometrics models task-level performance from task and system features~\citep{ge2026psychometrics}. ERB evaluates an agent's ability to choose and develop measurement strategies through autonomous, budgeted experimentation.

\subsection{Autonomous Research and Self-Improvement}
\label{sec:related-research}
Research benchmarks study iterative experimentation~\citep{huang2023mlagentbench,nathani2025mlgym}, resource-bounded machine-learning research~\citep{chan2024mlebench,wijk2024rebench}, replication~\citep{starace2025paperbench}, and post-training~\citep{rank2026posttrainbench}. The AI Scientist and Agent Laboratory integrate multiple stages of scientific work~\citep{lu2024aiscientist,schmidgall2025agentlaboratory}. Self-improvement systems such as FunSearch, AlphaEvolve, and the Darwin G\"odel Machine use evaluation feedback to guide program or agent revisions~\citep{romeraparedes2024funsearch,novikov2025alphaevolve,zhang2025dgm}; MIRROR and TRIAGE study self-knowledge and resource allocation~\citep{wang2026mirror,nazi2026triage}. ERB focuses on constructing the feedback instrument itself. Following artifact-based research benchmarks, it assesses the submitted evaluation independently of the process that produced it.

%% file: sections/method.tex
\section{EvalResearchBench}
\label{sec:method}

\subsection{Research Task}

\paragraph{Researcher, candidates, and targets.}
The \emph{researcher} is an LLM agent equipped with a harness that lets it inspect files, write and execute code, call models, and revise its work. A \emph{candidate} is a model--execution configuration whose capabilities the researcher must measure. A \emph{target benchmark} provides the reference candidate scores that the resulting evaluation should reproduce.

We cover three complementary capabilities needed for autonomous research. \emph{Coding} supports implementing and testing systems, from repository engineering to scientific programming and terminal workflows~\citep{deepswe2026,scicode2024,terminalbench2026}. \emph{Co-work} covers gathering information, following constraints, and producing professional deliverables~\citep{advancedif2025,browsecomp2025,gdpval2025}. \emph{Reasoning} covers analytical work with academic questions, long documents, and specialized physics~\citep{hle2025,aalcr2025,critpt2025}. A single research episode designs an evaluation across all three domains, allowing the agent to share resources and choose where further testing is most useful.

\paragraph{Input materials.}
We provide target descriptions, available original tasks and grading resources, historical execution traces, and reference scores for development candidates. The researcher also receives metered candidate-model APIs, an execution SDK, and the submission contract. Resource availability varies by target; the agent may select existing tasks, synthesize new ones, or combine both approaches. Development tests return candidate responses and grading results for diagnosing errors and improving correlation with the visible target references.

\paragraph{Budgets.}
We constrain three costs: researcher inference pays for the agent's analysis and tool use; development testing pays for candidate and helper-model calls; independent assessment pays to execute the frozen design. Keeping the first two budgets separate prevents agent deliberation from consuming the testing allowance. A fixed deadline limits research time. Section~\ref{sec:experimental-setup} gives the numerical limits.

The researcher may select, synthesize, or simplify tasks, repair graders, change task weights, set answer allowances, and inspect pilot responses. Calls and retries pass through a metered gateway. The agent allocates its development budget across exploration, testing, and revision without a prescribed sequence.

\paragraph{Submission.}
To require a reusable test suite rather than a few demonstrations, the agent submits at least 100 distinct scored questions, together with executable graders and fixed aggregation rules. The program returns one measured score per domain from candidate responses and frozen assets, without candidate-identity lookup or access to reference labels during scoring. Questions, grading code, and weights are frozen before assessment. One ERB entry is one research episode by a particular model and harness.

\subsection{Independent Assessment}

To test transfer beyond development feedback, we execute the frozen program independently on development and withheld candidates and compare its scores with visible and withheld target references. Candidate access and target-reference access are assigned separately (Table~\ref{tab:setup}).

The executor records per-item rewards, termination reasons, and usage. A total assessment cap limits the cost of evaluating the candidate panel, while a per-candidate cap bounds the cost of applying the design to any one model. Assessment is charged separately from development, and its results are not returned to the researcher for further revision.

A graded response is complete under the submitted contract, including a wrong or empty answer at the output cap. Infrastructure failures and budget-stopped items remain incomplete. Let $C_{a,m,d}$ be the completed items for researcher $a$, candidate $m$, and domain $d$, where $s_{a,m,i}$ is the item grade and $w_{a,i}\geq 0$ is the weight assigned by the researcher to question $i$. Weights specify each question's contribution to the domain score and are fixed before assessment. Its measured score is
\begin{equation}
 x_{a,m,d}=\frac{\sum_{i\in C_{a,m,d}}w_{a,i}s_{a,m,i}}{\sum_{i\in C_{a,m,d}}w_{a,i}}.
\end{equation}
A domain with no completed positive-weight item is missing. Coverage accompanies partial executions. Recorded expenditure separates researcher inference, development testing, and assessment, with unresolved reservations accounted for separately.

\subsection{Score Agreement and Ordering}
\label{sec:quality}

We assess both reference score levels and candidate ordering. We define \emph{Score} as 100 times the concordance correlation coefficient (CCC)~\citep{lin1989concordance}; pairwise agreement (PA) measures ordering.

For target $t$, match candidates with both an observed score $x_{a,m,d(t)}$ and a reference $y_{m,t}$. An eligible target has at least six matches and nonconstant references. Measured scores keep their native $[0,1]$ grading scale. Percentage references are divided by 100; Elo references use the minimum and range of the fixed reference panel. These constants are shared across researchers (Appendix~\ref{app:matched-comparison}). Each submitted domain score is compared unchanged with every eligible target in that domain.

Let $\bar x$, $\bar y$, $\operatorname{Var}(x)$, $\operatorname{Var}(y)$, and $\operatorname{Cov}(x,y)$ be population moments over matched candidates. Then
\begin{equation}
 \mathrm{Score}_{a,t}=100\frac{2\operatorname{Cov}(x,y)}{\operatorname{Var}(x)+\operatorname{Var}(y)+(\bar x-\bar y)^2},
 \qquad \mathrm{Score}_{a,\mathcal G}=\frac{1}{|\mathcal G|}\sum_{t\in\mathcal G}\mathrm{Score}_{a,t}.
\end{equation}
Exact agreement gives Score 100; constant measurements give zero, and opposing candidate differences can give negative values. CCC combines correlation with a penalty for mean and scale mismatch. Appendix~\ref{app:reference-compatibility} analyzes compatibility among the target score scales.

For the non-tied reference pairs $\mathcal P_t=\{(i,j):i<j,\ y_{i,t}\ne y_{j,t}\}$ among matched candidates, define
\begin{equation}
 \mathrm{PA}_t=\frac{1}{|\mathcal P_t|}\sum_{(i,j)\in\mathcal P_t}
 \left[\mathbf{1}\!\left\{(x_{i,d}-x_{j,d})(y_{i,t}-y_{j,t})>0\right\}
 +\tfrac12\mathbf{1}\!\left\{x_{i,d}=x_{j,d}\right\}\right].
\end{equation}
Correct order receives one point, a measured tie half a point, and a reversal zero. Reference ties are excluded; no significance threshold is applied. Both metrics average targets equally within each domain or visibility group; overall averages all eligible targets.

%% file: sections/experiments.tex
\section{Experiments}
\label{sec:experiments}

\subsection{Setup}
\label{sec:experimental-setup}

We compare how different LLM researchers use the available materials and budget to build evaluations. The experiments first assess the resulting measurements, then examine the design choices and costs behind them. Appendix~\ref{app:research-case} traces how these choices arise during research.

\paragraph{Models.}
We study nine researcher configurations: GPT-6 Astra~\citep{openai2026gpt6astra}, GPT-5.6 Sol~\citep{openai2026gpt56}, GLM-5.3~\citep{zai2026glm53}, DeepSeek V4 Pro~\citep{deepseek2026v4}, Kimi K3~\citep{moonshot2026kimik3}, Fable 5.1~\citep{anthropic2026fable51}, Opus 5~\citep{anthropic2026opus5}, Grok 4.7~\citep{xai2026grok47}, and Fable 5~\citep{anthropic2026fable5}. Each contributes one research episode. The GPT researchers and Grok use Codex~\citep{openai2025codex}; the other researchers use Claude Code~\citep{anthropic2025claudecode}. The candidate panel contains four development-visible and nine sealed models (Table~\ref{tab:setup}).
\input{figures/table_setup}

\paragraph{Budgets and implementation.}
The research budget is three hours, \$100 for researcher inference, and \$100 for shared development/helper testing. Independent assessment has a \$120 total cap and a \$30 per-candidate cap. Pilots may use smaller suites than the final submission; the researcher allocates its development calls freely.

\paragraph{Public-task sampling baseline.}
A research agent should be compared with a simpler way to build a broad evaluation: sample questions from existing public sources. We use \emph{Random120}, a source-stratified baseline that fixes the number of questions drawn from each source, ensuring coverage across task types before sampling. Specifically, we select sixteen public task sources, eligibility filters, source quotas, local graders, and equal item weighting within each domain. A seeded draw within each source selects 40 coding, 40 co-work, and 40 reasoning tasks without using target reference scores. We assess its twelve complete candidates against the same fourteen targets (Appendix~\ref{app:random120-baseline}).

\subsection{Main Results: Agreement with the Reference Panel}
\label{sec:main-results}
\input{figures/table_main_results}

Eight submissions yield all 39 candidate-domain scores. Fable 5 reaches its assessment budget before completing the panel; Table~\ref{tab:main-results} reports its stated imputation scenario separately from the eight observed-score results.

\paragraph{Overall results and domain strengths.}
Astra leads overall Score, whereas Grok leads PA: reproducing score levels and recovering candidate order favor different designs. Domain strengths also vary, with no researcher dominating both metrics across all three domains. Decomposing CCC as $rC_b$ helps explain why: Astra and GLM-5.3 have similar co-work correlations, but Astra aligns more closely with the reference scale (Table~\ref{tab:concordance-factors}). A useful ordering alone does not guarantee faithful score reproduction.

\paragraph{Development and sealed targets reveal different strengths.}
Fable 5.1 leads development Score, but Grok leads sealed Score. This reversal makes cross-target transfer a distinct requirement for evaluation research: a useful design must preserve capability signals beyond the references used to develop it. Systematic score shifts across targets within the same domain (Figure~\ref{fig:family-scatter}) further show why a strong fit to one benchmark does not establish a broadly useful evaluator.

\begin{figure}[!htb]
\centering
\includegraphics[width=.90\linewidth]{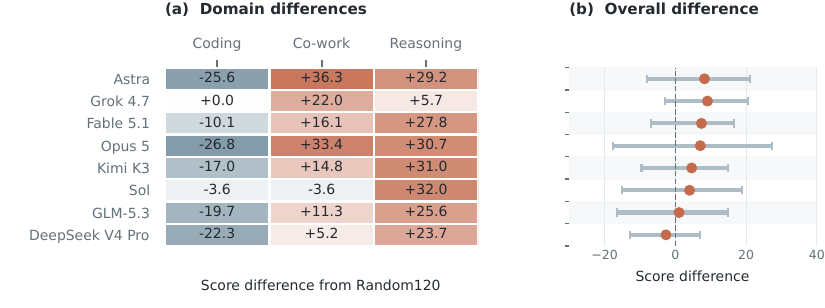}
\caption{\textbf{Where research differs from fixed public-task sampling.} Differences from Random120 on identical candidate--target intersections: (a) per domain; (b) overall, with central 95\% candidate-resampling ranges. The twelve candidates are jointly resampled; fourteen targets and frozen evaluations stay fixed.}
\label{fig:matched-baseline}
\end{figure}

\paragraph{Public-task sampling is a strong baseline.}
Random120's public sources, domain quotas, and graders encode task-design knowledge, making it a strong baseline, especially in coding. Its coding and reasoning coverage recovers much of the candidate ordering, while Astra's Score gains concentrate in co-work and reasoning (Figure~\ref{fig:matched-baseline}). DeepSeek V4 Pro illustrates why iteration alone need not beat this baseline. It revises coding tasks and increases reasoning allowances, yet its short, deterministic coding and co-work checks yield scores clustered near the ceiling; its reasoning gains cannot offset the coding deficit. Research adds value when it identifies tasks that preserve target-relevant differences between candidates. Ease of grading and repeated revision do not by themselves ensure informative measurements.

%% file: figures/table_setup.tex
\begin{table}[!htbp]
\centering
\caption{\textcolor{tableink}{\textbf{Experimental setup.}} Three domains, thirteen candidates, and fourteen targets. Candidate and target visibility are independent; reference scores are fixed public snapshots.}
\label{tab:setup}
\scriptsize
\setlength{\tabcolsep}{4pt}
\renewcommand{\arraystretch}{1.08}
\begin{tabularx}{\textwidth}{@{}>{\raggedright\arraybackslash}p{0.15\textwidth}>{\raggedright\arraybackslash}X>{\raggedright\arraybackslash}X@{}}
\toprule
\textcolor{tableink}{\textbf{Component}} & \textcolor{tableaccent}{\textbf{Development-visible}} & \textcolor{tableaccent}{\textbf{Sealed}} \\
\midrule
Candidates & GPT-5.5; GLM-5.2; DeepSeek V4 Flash-0731; Qwen3.8 Max (4)
& GPT-6 Astra; GPT-5.6 Sol; Kimi K3; MiMo V2.5 Pro; DeepSeek V3.2; Gemma 4 31B; Qwen3.5 397B-A17B; GPT-OSS 120B; Nemotron 3 Ultra (9) \\
\midrule
Coding & DeepSWE; SciCode; SWE-bench Multilingual (3)
& Terminal-Bench 2.1; ProgramBench (Vals); SWE-bench (Vals) (3) \\
Co-work & GDPval; AdvancedIF (2)
& BrowseComp; Harvey LAB-AA; AA-Briefcase (3) \\
Reasoning & HLE (1)
& AA-LCR; CritPt (2) \\
\midrule
Research access & Candidate APIs, target descriptions, available development references, permitted tasks and grading resources
& Candidate APIs and target references withheld from the researcher until after submission freeze \\
\bottomrule
\end{tabularx}
\end{table}

%% file: figures/table_main_results.tex
\input{figures/table_full_results}

%% file: figures/table_full_results.tex
\begin{table}[!htbp]
\centering
\caption{\textbf{Frozen evaluation scores on development, sealed, and all targets.} Each group reports target-equal Score (100 times CCC) and PA (\%) for coding, co-work, reasoning, and their target-equal average. Rows are ordered by overall Score. \textbf{Bold}/\underline{underlined} mark the highest/second-highest observed values among the eight assessed artifacts. Random120 uses twelve complete candidates; the artifacts use thirteen. Fable 5 is a separate sensitivity scenario assigning 19 missing domain scores 100\%.}
\label{tab:main-results}
\label{tab:full-results}
\footnotesize
\setlength{\tabcolsep}{1.5pt}
\renewcommand{\arraystretch}{1.15}
\resizebox{\textwidth}{!}{%
\begin{tabular}{llrrrr|rrrr|rrrr}
\toprule
 & & \multicolumn{4}{c|}{Development (6 targets)} & \multicolumn{4}{c|}{Sealed (8 targets)} & \multicolumn{4}{c}{Overall (14 targets)} \\
\cmidrule(lr){3-6}\cmidrule(lr){7-10}\cmidrule(lr){11-14}
Evaluator & Metric & Coding & Co-work & Reasoning & Avg. & Coding & Co-work & Reasoning & Avg. & Coding & Co-work & Reasoning & Avg. \\
\midrule
GPT-6 Astra & Score$\uparrow$ & -5.19 & \textbf{40.41} & 60.64 & \underline{20.98} & 4.79 & \underline{43.16} & 24.69 & 24.15 & -0.20 & \textbf{42.06} & 36.67 & \textbf{22.79} \\
 & PA$\uparrow$ & 55.71 & \textbf{75.76} & \textbf{83.33} & 66.99 & 60.32 & 79.47 & \textbf{76.09} & 71.44 & 58.01 & \underline{77.98} & \textbf{78.51} & 69.54 \\
\addlinespace[2pt]
Grok 4.7 & Score$\uparrow$ & \textbf{24.35} & 10.34 & 13.33 & 17.84 & \textbf{22.22} & 37.84 & 14.48 & \textbf{26.14} & \textbf{23.28} & 26.84 & 14.10 & \underline{22.59} \\
 & PA$\uparrow$ & \textbf{77.49} & 72.10 & \underline{82.22} & \textbf{76.48} & \underline{66.85} & \underline{81.63} & 71.93 & \underline{73.66} & \textbf{72.17} & 77.81 & 75.36 & \textbf{74.87} \\
\addlinespace[2pt]
Fable 5.1 & Score$\uparrow$ & 9.81 & 23.52 & \underline{63.94} & \textbf{23.40} & \underline{20.92} & 19.68 & 22.60 & 20.87 & 15.36 & 21.22 & 36.38 & 21.96 \\
 & PA$\uparrow$ & 61.03 & 69.87 & 81.11 & 67.32 & 63.94 & 73.13 & 67.05 & 68.16 & 62.48 & 71.82 & 71.73 & 67.80 \\
\addlinespace[2pt]
Opus 5 & Score$\uparrow$ & -6.36 & \underline{24.34} & \textbf{77.99} & 17.93 & 5.75 & \textbf{47.60} & 18.70 & \underline{24.68} & -0.31 & \underline{38.30} & \textbf{38.46} & 21.79 \\
 & PA$\uparrow$ & 51.26 & 69.02 & 80.00 & 61.97 & 56.30 & 74.61 & 71.89 & 67.06 & 53.78 & 72.37 & 74.59 & 64.88 \\
\addlinespace[2pt]
Kimi K3 & Score$\uparrow$ & 6.49 & 12.34 & 55.33 & 16.58 & 7.93 & 24.16 & \textbf{28.35} & 19.12 & 7.21 & 19.43 & \underline{37.35} & 18.03 \\
 & PA$\uparrow$ & 64.33 & \underline{74.07} & 81.11 & \underline{70.38} & \textbf{70.88} & \textbf{87.07} & 73.99 & \textbf{77.73} & \underline{67.61} & \textbf{81.87} & 76.36 & \underline{74.58} \\
\addlinespace[2pt]
GPT-5.6 Sol & Score$\uparrow$ & \underline{21.90} & -6.66 & 50.45 & 17.14 & 17.71 & 10.24 & \underline{25.58} & 16.88 & \underline{19.81} & 3.48 & 33.87 & 16.99 \\
 & PA$\uparrow$ & \underline{66.94} & 46.38 & 81.11 & 62.45 & 60.60 & 60.31 & \underline{74.54} & 63.98 & 63.77 & 54.74 & \underline{76.73} & 63.32 \\
\addlinespace[2pt]
GLM-5.3 & Score$\uparrow$ & 1.86 & 11.68 & 50.00 & 13.15 & 8.73 & 18.30 & 19.91 & 15.12 & 5.29 & 15.65 & 29.94 & 14.27 \\
 & PA$\uparrow$ & 56.67 & 67.51 & 72.22 & 62.87 & 53.24 & 77.57 & 68.18 & 66.10 & 54.95 & 73.55 & 69.53 & 64.72 \\
\addlinespace[2pt]
DeepSeek V4 Pro & Score$\uparrow$ & 3.20 & 5.55 & 44.00 & 10.78 & -0.53 & 11.59 & 24.51 & 10.28 & 1.33 & 9.17 & 31.01 & 10.49 \\
 & PA$\uparrow$ & 56.78 & 64.27 & 74.44 & 62.22 & 63.19 & 65.91 & 64.18 & 64.46 & 59.99 & 65.25 & 67.60 & 63.50 \\
\addlinespace[2pt]
\midrule
Fable 5 (imputed) & Score$\uparrow$ & -7.77 & -12.58 & -18.63 & -11.18 & -5.68 & -21.07 & -18.58 & -14.68 & -6.72 & -17.67 & -18.60 & -13.18 \\
 & PA$\uparrow$ & 44.42 & 44.78 & 37.78 & 43.43 & 53.47 & 38.13 & 26.81 & 41.06 & 48.95 & 40.79 & 30.47 & 42.07 \\
Random120 & Score$\uparrow$ & 21.44 & -10.71 & 12.79 & 9.28 & 24.39 & 13.39 & 7.15 & 15.96 & 22.92 & 3.75 & 9.03 & 13.10 \\
 & PA$\uparrow$ & 73.87 & 55.96 & 68.06 & 66.93 & 76.63 & 62.06 & 69.20 & 69.31 & 75.25 & 59.62 & 68.82 & 68.29 \\
\bottomrule
\end{tabular}}
\end{table}

%% file: sections/agent_behavior_analysis.tex
\subsection{What Shapes Evaluation Quality?}
\label{sec:design-analysis}

We analyze three design decisions---task selection, question count, and aggregation---by varying one part of each evaluation while holding candidate responses and references fixed.

\begin{figure}[!htb]
\centering
\includegraphics[width=\linewidth]{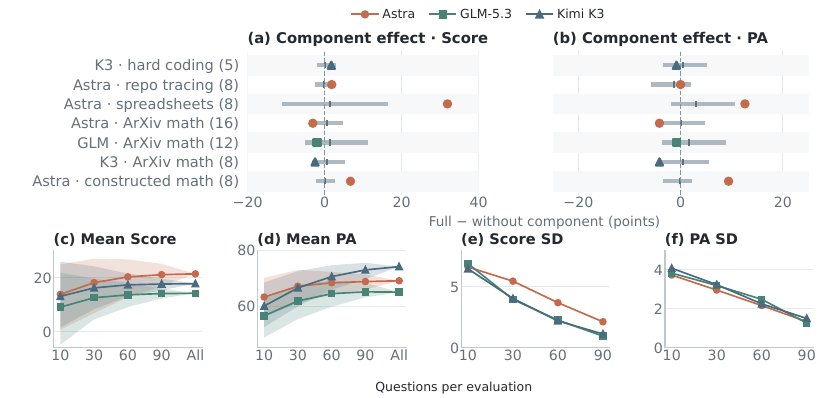}
\caption{\textbf{Task contribution and measurement variability.} (a,b) Full-minus-deleted Score and PA for seven task groups; parentheses give question counts. Gray intervals cover the central 95\% of 1,000 same-domain, same-size random deletions. (c,d) Mean overall Score and PA across nested question subsets, with central 95\% bands. (e,f) Their standard deviations. Candidates, targets, and responses are fixed.}
\label{fig:research-diagnostics}
\end{figure}

\paragraph{Choose tasks for their measurement contribution.}
K3 adds hard coding problems after near-ceiling pilots; Astra adds repository traces after inspecting upstream software tasks. Removing these groups, however, has only modest effects relative to same-domain, same-size random deletions. K3's addition improves Score but slightly reduces PA; Astra's repository group leaves PA unchanged. By contrast, Astra's spreadsheet tasks contribute strongly to both co-work metrics (Figure~\ref{fig:research-diagnostics}a,b). Task realism and difficulty motivate a design, but its value depends on how it separates candidates in relation to the target.

Many Astra spreadsheet responses exhaust their answer allowance, so the observed separation reflects the task under its chosen execution conditions. Grok responds to truncated hard-task pilots by favoring shorter, self-contained checks and adjusting answer allowances. DeepSeek instead expands reasoning allowances and selects short exact-answer items. A short expected answer does not itself make a problem easy to solve.

\input{sections/ablations}

\begin{figure}[!htb]
\centering
\includegraphics[width=.81\linewidth]{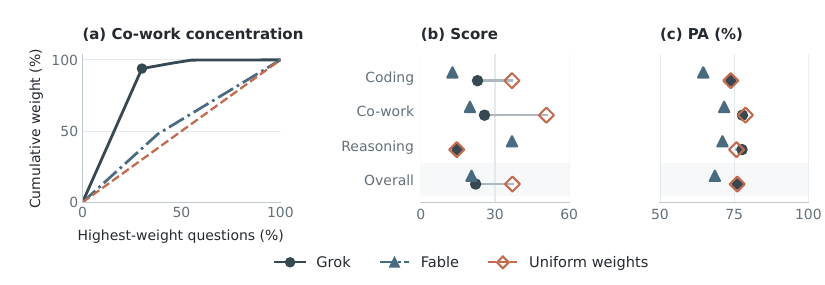}
\caption{\textbf{Weight allocation and measurement quality.} Grok and Fable (Fable 5.1) use their submitted weights. (a) Cumulative co-work weight versus the fraction of questions, sorted by decreasing weight; the dashed diagonal denotes uniform weighting. (b,c) Score and PA on the same twelve-candidate, fourteen-target panel. Here \emph{Uniform weights} reweights Grok's grades only; connectors show this change, with its responses and item exclusions fixed.}
\label{fig:grok-weights}
\end{figure}

\paragraph{Use weights to preserve more than a development ordering.}
Grok groups pilot questions by which development candidates pass, then searches for weights that reproduce the visible ordering and score gaps. For co-work, it assigns most weight to questions that all three tested development candidates pass, using a small remainder to shape their differences. Consequently, six of its twenty co-work questions carry almost all scoring weight. Fable 5.1 distributes influence more evenly, with a modest upweighting of professional tasks (Figure~\ref{fig:grok-weights}a).

Recomputing Grok's scores with equal item weights on the same responses substantially improves coding and co-work Score, while overall PA barely changes (Figure~\ref{fig:grok-weights}). The Score improvement also holds on the complete-item candidate subset. The frozen suite therefore contains useful measurement signal that its chosen weights underuse. Evaluation design must consider task contribution, sampling variability, and score aggregation jointly.

\subsection{What Makes an Evaluation Cost-Efficient?}
\label{sec:cost-analysis}

\paragraph{Low execution cost can preserve measurement quality.}
An evaluator is built once but may be run repeatedly, so we separate construction cost from frozen assessment cost. Grok combines low assessment cost with the highest overall PA and a competitive Score (Figure~\ref{fig:cost-analysis}; Table~\ref{tab:main-results}). Grok and Fable 5.1 submit similar question counts, yet their execution costs differ substantially. Task count alone therefore gives little indication of an evaluator's recurring cost.
\input{figures/figure_cost_analysis}

\paragraph{Pilot feedback shapes answering and grading costs.}
Grok's hard HLE pilots exhaust their answer allowance without producing visible answers. It responds by favoring short code, output predictions, and numeric checks with local graders. Fable 5.1 retains longer professional deliverables and increases their answer allowance after truncated pilots; its rubric judge adds a substantial second expense. Opus incurs little grading expense but still spends heavily on candidate answers. These trajectories identify two distinct design choices behind execution cost: what a candidate must produce and how its answer is judged (Appendix~\ref{app:cost-design}).

\section{Case Study: How Researchers Revise Evaluations}
\label{sec:case-study}
\label{sec:behavior-analysis}

Researchers use development feedback to expand coverage, repair tasks, and adjust execution. Astra develops repository-tracing tasks from source inspection, resolves prompt--rubric mismatches through pilot feedback, and adjusts execution after truncated responses. GLM corrects invalid examples and numerical answers, while K3 adds harder coding tasks after near-ceiling pilot scores. Section~\ref{sec:design-analysis} analyzes these design choices; Appendix~\ref{app:research-case} details the cases.

%% file: sections/ablations.tex
\paragraph{More questions improve both average quality and consistency.}
\label{sec:ablations}

We draw nested, domain-stratified subsets from the frozen Astra, GLM-5.3, and K3 suites and recompute their scores on the same candidate panel. As subsets grow, mean Score rises and both Score and PA fluctuate less across question draws (Figure~\ref{fig:research-diagnostics}c--f). Small subsets can omit useful task families or overrepresent a few responses, even when their domain proportions are fixed. The design problem is to retain this signal and consistency with fewer calls.

%% file: figures/figure_cost_analysis.tex
\begin{figure}[!htb]
\centering
\includegraphics[width=.81\linewidth]{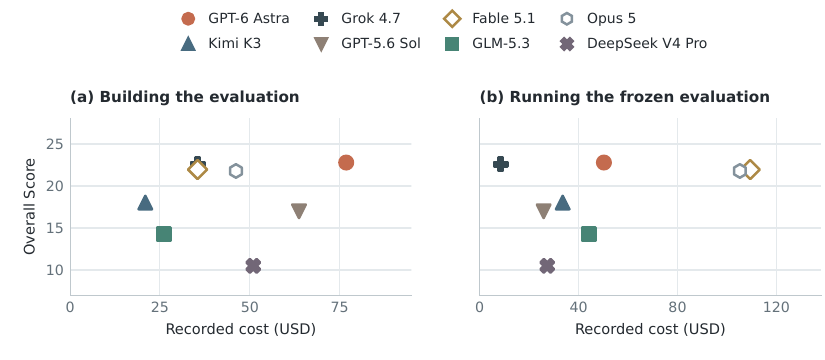}
\caption{\textbf{Spending and measurement quality.} Overall Score against recorded API-equivalent costs for (a) researcher inference plus development tests and (b) frozen assessment. Solid markers use metered usage; hollow markers use collaborator reports. Unresolved reservations are excluded.}
\label{fig:cost-analysis}
\end{figure}

%% file: sections/conclusion.tex
\section{Conclusion}
\label{sec:conclusion}
ERB measures autonomous evaluation research under fixed budgets. Independent assessment separates score reproduction, ordering, and execution cost. Source inspection and pilot audits reveal how agents revise tasks and graders; component and subset analyses show why difficulty, nominal size, and measurement value must be assessed separately. This establishes an executable benchmark for the evaluation-design capability that autonomous improvement requires.

%% file: sections/appendix_comparison.tex
\section{Target-Level Assessment and Score Scales}
\label{app:matched-comparison}

Table~\ref{tab:main-results} presents the complete development, sealed, and overall breakdown in the main text. This appendix gives target-level results and the reference-scale conventions.

Assessment tables use Score and PA from Section~\ref{sec:quality}, with target-equal aggregation and the same reference-bearing candidate IDs per target. Domain columns average their own targets. Frozen-component and question-count analyses use these same two metrics on a common twelve-candidate panel.

Measured scores stay on their frozen $[0,1]$ grading scale. Percentage references use offset zero and divisor 100, including GDPval, Harvey criterion pass rate, ProgramBench strict resolution, and CritPt. AA-Briefcase is an Elo reference and uses the following fixed panel constants, shared by all artifacts:
\begin{center}\begin{tabular}{lrr}\toprule
Reference & Offset & Divisor\\\midrule
AA-Briefcase Elo & 0.00 & 1640.35 \\
\bottomrule\end{tabular}\end{center}
The Elo constants are the minimum and range over the frozen reference-bearing candidate panel, shared across all researchers. This convention fixes the unit of score differences for the panel.

The development group contains DeepSWE~\citep{deepswe2026}, SciCode~\citep{scicode2024}, SWE-bench Multilingual~\citep{yang2025swesmith}, AdvancedIF~\citep{advancedif2025}, GDPval~\citep{gdpval2025}, and HLE~\citep{hle2025}. The sealed group contains Terminal-Bench 2.1~\citep{terminalbench2026}, BrowseComp~\citep{browsecomp2025}, AA-LCR~\citep{aalcr2025}, ProgramBench (Vals)~\citep{yang2026programbench}, SWE-bench (Vals)~\citep{jimenez2024swebench}, Harvey LAB-AA~\citep{harvey2026lab}, AA-Briefcase~\citep{artificialanalysis2026briefcase}, and CritPt~\citep{critpt2025}. The criterion comprises published score snapshots from their source protocols. Nine reference sets come from Mercor's APEX open-source leaderboards~\citep{mercor2026apex}: DeepSWE, SciCode, SWE-bench Multilingual, Terminal-Bench 2.1, AdvancedIF, GDPval, BrowseComp, HLE, and AA-LCR. ProgramBench (Vals) and SWE-bench (Vals) come from Vals AI~\citep{vals2026programbench,vals2026swebench}, and Harvey LAB-AA, AA-Briefcase, and CritPt from Artificial Analysis~\citep{artificialanalysis2026harveylab,artificialanalysis2026briefcaseleaderboard,artificialanalysis2025critpt}.

The four development-visible candidates are GPT-5.5~\citep{openai2026gpt55}, GLM-5.2~\citep{zai2026glm52}, DeepSeek V4 Flash-0731~\citep{deepseek2026v4}, and Qwen3.8 Max~\citep{qwen2026qwen38max}. The nine sealed candidates are GPT-6 Astra~\citep{openai2026gpt6astra}, GPT-5.6 Sol~\citep{openai2026gpt56}, Kimi K3~\citep{moonshot2026kimik3}, MiMo V2.5 Pro~\citep{xiaomi2026mimov25pro}, DeepSeek V3.2~\citep{deepseek2025v32}, Gemma 4 31B~\citep{google2026gemma4}, Qwen3.5 397B-A17B~\citep{qwen2026qwen35}, GPT-OSS 120B~\citep{openai2025gptoss}, and Nemotron 3 Ultra~\citep{nvidia2026nemotron3ultra}.
\subsection{Representative target scatterplots}
Figure~\ref{fig:family-scatter} shows one development and one sealed target in each domain for Astra, K3, and Fable 5.1. Astra obtains strong co-work agreement on both GDPval (Score 55.78) and BrowseComp (42.23). In reasoning, the three evaluators reach Scores of 55.33--63.94 on HLE, but only 3.62--7.83 on AA-LCR, where their measured scores lie systematically below the references. A single domain score can therefore preserve different aspects of different targets.
\input{figures/reference_scatter_figure}

\paragraph{Incompatibility of two coding references.}
\label{app:reference-compatibility}
A single domain score is compared with several target references, which need not agree in level or ordering. We first quantify how this disagreement limits raw-score reproduction. ProgramBench (Vals) and SWE-bench Multilingual share six reference-bearing candidates. Let $y_{m,1}$ and $y_{m,2}$ be their fixed-scale reference scores and $x_m$ a single coding score for candidate $m$. The midpoint $x_m=(y_{m,1}+y_{m,2})/2$ minimizes pooled squared error to both references, giving
\begin{equation}
\min_{x_1,\ldots,x_6}\sqrt{\frac{1}{12}\sum_{m=1}^{6}\sum_{j=1}^{2}(x_m-y_{m,j})^2}
=\frac{1}{2}\sqrt{\frac{1}{6}\sum_{m=1}^{6}(y_{m,1}-y_{m,2})^2}=0.4356.
\end{equation}
The reference means are 0.0075 and 0.876 on the same six candidates. The resulting raw-score error floor belongs to the reference pair; CCC also depends on covariance and dispersion. PA separately reports ordering.

The same calculation extends to every reference in a domain. For target set $T_d$, minimize the target-equal mean squared error $|T_d|^{-1}\sum_{t\in T_d}|\mathcal M_t|^{-1}\sum_{m\in\mathcal M_t}(x_m-y_{m,t})^2$ over one common $x_m\in[0,1]$ per candidate. Its exact minimizer is the weighted mean of that candidate's available reference scores, with target $t$ weighted by $1/|\mathcal M_t|$. The resulting oracle RMSE floors are:
\begin{center}
\begin{tabular}{lrrr}
\toprule
Domain & Development & Sealed & Overall \\
\midrule
Coding & 0.209 & 0.353 & 0.303 \\
Co-work & 0.074 & 0.088 & 0.128 \\
Reasoning & 0.000 & 0.284 & 0.244 \\
\bottomrule
\end{tabular}
\end{center}
These label-informed values characterize the reference panel's raw-score compatibility. The zero for single-target development reasoning follows from assigning its reference score directly; the RMSE floors are distinct from CCC ceilings.

\paragraph{Compatible-reference sensitivity.}
To examine how reference-scale differences affect the results, we form an exploratory target subset without reading any researcher score: in each domain, choose the largest group in which every pair shares at least six reference-bearing candidates and has reference-score RMSE at most 0.25; break ties by the group's mean pairwise RMSE. This selects SWE-bench Multilingual, SWE-bench (Vals), and Terminal-Bench 2.1 for coding; AdvancedIF, BrowseComp, and GDPval for co-work; and HLE and CritPt for reasoning. Four selected targets are development-visible and four are sealed. Target-equal Score and PA on this eight-target subset are:
\begin{center}
\begin{tabular}{lrr}
\toprule
Frozen artifact & Score $\uparrow$ & PA (\%) $\uparrow$ \\
\midrule
Fable 5.1 & 33.47 & 69.57 \\
Astra & 27.51 & 67.08 \\
Grok & 23.95 & 75.06 \\
Sol & 22.31 & 59.82 \\
Opus 5 & 22.29 & 62.73 \\
K3 & 22.12 & 73.08 \\
DeepSeek V4 Pro & 17.46 & 65.00 \\
GLM-5.3 & 17.32 & 61.28 \\
\bottomrule
\end{tabular}
\end{center}
This scale-based sensitivity checks how target composition affects raw-score concordance. Fable 5.1 leads the eight-target subset, while Astra leads the primary fourteen-target panel.

\paragraph{Reference-imposed ordering limit.}
Reference targets in one domain sometimes order the same candidate pair differently. We calculate the largest PA attainable by \emph{any} single strict candidate ordering per domain on the fixed panel. Every target and every non-tied pair within it receives exactly the weight used in Section~\ref{sec:quality}; an exact subset dynamic program maximizes the weighted agreement. A measured tie earns the average of its two possible orientations and cannot improve the optimum. This label-informed oracle is an upper bound for a frozen evaluator, not a deployable score:
\begin{center}
\begin{tabular}{lrrr}
\toprule
Domain & Development & Sealed & Overall \\
\midrule
Coding & 92.11\% & 94.72\% & 92.36\% \\
Co-work & 84.68\% & 95.40\% & 88.72\% \\
Reasoning & 100.00\% & 93.18\% & 93.03\% \\
All targets & 90.95\% & 94.59\% & 91.20\% \\
\bottomrule
\end{tabular}
\end{center}
Each column optimizes its own target set, so the overall maximum need not average the development and sealed maxima. Cross-target order conflicts cap overall PA at 91.20\%, compared with the best observed artifact's 74.87\%.

Fable 5 produced 20 of 39 candidate-domain reports before assessment stopped at the budget cap. The main ranking excludes it. Table~\ref{tab:main-results} separately reports the fixed scenario that assigns its 19 unobserved domain scores a value of one; these are scenario-conditional metrics, not observed-panel estimates or metric bounds.

\paragraph{Fixed-score sanity check.}
To check whether an uninformative evaluator can obtain a favorable metric, we assign every candidate a score of $0.5$ in every domain without making model calls. This yields overall Score$=0$ and PA$=50.00\%$ on the same fourteen target panels. All measured pairs tie. Score requires score variation aligned with the reference and penalizes mean and scale mismatch.

\input{figures/table_target_results}

\subsection{Cost accounting}
\label{app:cost-accounting}
Costs separate researcher inference, shared development/helper testing, and frozen assessment. The values are standardized API-equivalent estimates.

\input{sections/appendix_cost_design}

%% file: figures/reference_scatter_figure.tex
\begin{figure}[!htb]
\centering
\includegraphics[width=\textwidth]{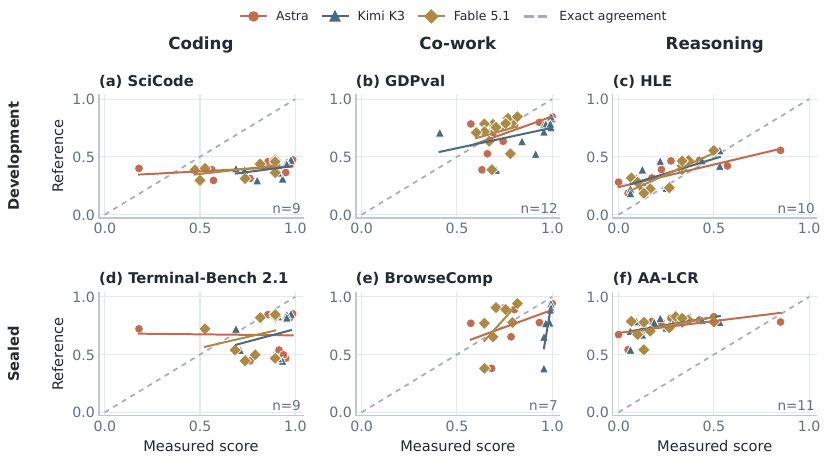}
\caption{\textbf{Measured scores versus target references.} Columns show coding, co-work, and reasoning; rows show one development and one sealed target per domain. Points denote candidates, colored lines show within-panel OLS trends, and the dashed line marks exact agreement. Score and PA are computed from the original points.}
\label{fig:family-scatter}
\end{figure}

%% file: figures/table_target_results.tex
\begin{table*}[t]\centering\scriptsize
\setlength{\tabcolsep}{3pt}
\begin{tabular}{llllrrrrrr}\toprule
Target & Group & $n$ & Metric & Astra & Grok & Fable 5.1 & Opus 5 & K3 & Sol \\ \midrule
DeepSWE & Dev & 9 & PA$\uparrow$ & 53.03 & 84.85 & 59.09 & 50.00 & 63.64 & 78.79 \\
 &  &  & Score$\uparrow$ & -2.32 & 21.48 & 8.61 & -1.08 & 2.76 & 14.54 \\
SciCode & Dev & 9 & PA$\uparrow$ & 69.44 & 83.33 & 72.22 & 68.06 & 72.22 & 66.67 \\
 &  &  & Score$\uparrow$ & 5.73 & 2.75 & 7.54 & 7.39 & 1.80 & 2.15 \\
SWE-bench Multilingual & Dev & 8 & PA$\uparrow$ & 44.64 & 64.29 & 51.79 & 35.71 & 57.14 & 55.36 \\
 &  &  & Score$\uparrow$ & -18.99 & 48.81 & 13.27 & -25.37 & 14.91 & 49.01 \\
Terminal-Bench 2.1 & Sealed & 9 & PA$\uparrow$ & 55.56 & 75.00 & 66.67 & 50.00 & 63.89 & 58.33 \\
 &  &  & Score$\uparrow$ & -1.85 & 26.70 & 18.99 & -21.84 & 10.64 & 20.44 \\
ProgramBench (Vals) & Sealed & 8 & PA$\uparrow$ & 68.75 & 50.00 & 56.25 & 66.67 & 85.42 & 64.58 \\
 &  &  & Score$\uparrow$ & 0.53 & -0.04 & 0.32 & 0.59 & 0.24 & 0.12 \\
SWE-bench (Vals) & Sealed & 10 & PA$\uparrow$ & 56.67 & 75.56 & 68.89 & 52.22 & 63.33 & 58.89 \\
 &  &  & Score$\uparrow$ & 15.69 & 39.99 & 43.45 & 38.49 & 12.90 & 32.59 \\
\addlinespace[3pt]
AdvancedIF & Dev & 11 & PA$\uparrow$ & 66.67 & 63.89 & 68.52 & 59.26 & 64.81 & 35.19 \\
 &  &  & Score$\uparrow$ & 25.03 & -4.61 & 28.14 & -0.67 & -1.00 & -18.32 \\
GDPval & Dev & 12 & PA$\uparrow$ & 84.85 & 80.30 & 71.21 & 78.79 & 83.33 & 57.58 \\
 &  &  & Score$\uparrow$ & 55.78 & 25.28 & 18.90 & 49.36 & 25.68 & 5.01 \\
BrowseComp & Sealed & 7 & PA$\uparrow$ & 71.43 & 85.71 & 80.95 & 76.19 & 95.24 & 57.14 \\
 &  &  & Score$\uparrow$ & 42.23 & 32.32 & 39.52 & 28.61 & 6.06 & -8.04 \\
Harvey LAB-AA & Sealed & 9 & PA$\uparrow$ & 83.33 & 79.17 & 63.89 & 69.44 & 77.78 & 58.33 \\
 &  &  & Score$\uparrow$ & 42.54 & 32.26 & 3.72 & 58.76 & 31.02 & 10.50 \\
AA-Briefcase & Sealed & 11 & PA$\uparrow$ & 83.64 & 80.00 & 74.55 & 78.18 & 88.18 & 65.45 \\
 &  &  & Score$\uparrow$ & 44.73 & 48.95 & 15.81 & 55.43 & 35.40 & 28.24 \\
\addlinespace[3pt]
HLE & Dev & 10 & PA$\uparrow$ & 83.33 & 82.22 & 81.11 & 80.00 & 81.11 & 81.11 \\
 &  &  & Score$\uparrow$ & 60.64 & 13.33 & 63.94 & 77.99 & 55.33 & 50.45 \\
AA-LCR & Sealed & 11 & PA$\uparrow$ & 78.70 & 70.37 & 66.67 & 74.07 & 72.22 & 74.07 \\
 &  &  & Score$\uparrow$ & 7.83 & 19.18 & 3.62 & 5.67 & 4.29 & 3.80 \\
CritPt & Sealed & 12 & PA$\uparrow$ & 73.48 & 73.48 & 67.42 & 69.70 & 75.76 & 75.00 \\
 &  &  & Score$\uparrow$ & 41.55 & 9.78 & 41.58 & 31.73 & 52.42 & 47.36 \\
\addlinespace[3pt]
\bottomrule\end{tabular}
\caption{Per-target PA (\%) and Score for the four highest overall-Score artifacts, plus K3 and Sol. Scores use the frozen metrics and target panels in Section~\ref{sec:quality}; $n$ is the number of reference-bearing candidates.}
\label{tab:matched-targets}\label{tab:pairwise-targets}
\end{table*}

%% file: sections/appendix_cost_design.tex
\subsection{Task design, execution cost, and score concentration}
\label{app:cost-design}

We compare frozen task forms, scoring weights, and recorded execution cost for the suites discussed in Section~\ref{sec:cost-analysis}.

Table~\ref{tab:execution-cost-components} separates candidate generation from rubric and equivalence grading costs.
\input{figures/table_execution_cost_components}

\begin{table}[ht]
\centering
\small
\setlength{\tabcolsep}{5pt}
\renewcommand{\arraystretch}{1.13}
\begin{tabularx}{\linewidth}{@{}p{.13\linewidth}>{\raggedright\arraybackslash}X>{\raggedright\arraybackslash}X@{}}
\toprule
Dimension & Grok & Fable 5.1 \\
\midrule
Coding & 62 tasks: 39 short Python implementations and 23 program-output predictions. & 38 tasks: 20 SciCode and 18 LiveCodeBench~\citep{jain2024livecodebench}. \\
Co-work & 20 tasks: 12 numeric answers and 8 short format/constraint tasks. & 42 tasks: 26 Advanced-IF and 16 GDPval professional tasks. \\
Reasoning & 25 tasks: 3 short HLE multiple-choice questions and 22 short numeric/probability problems. & 30 HLE tasks. \\
Response form & Short function, program output, scalar/fraction, option letter, or constrained note. & Scientific/competitive code, instruction following, professional tasks, and HLE, verified in the frozen suite. \\
Scoring weights & Nonuniform: six co-work tasks carry 93.87\%; three HLE tasks carry 3.60\% of reasoning. & Coding/reasoning equal-weighted. Co-work IF weight 1 and GDPval weight 1.5. \\
\bottomrule
\end{tabularx}
\caption{\textbf{Task forms and weights at similar nominal scale.} Grok has 107 questions and Fable 5.1 has 110.}
\label{tab:cost-design}
\end{table}

Grok's short functions, program-output predictions, scalar answers, and format checks contrast with Fable 5.1's scientific code, professional tasks, and HLE questions. Grok's graders execute code or compare explicit answers and formats.

Grok's prompts have median length 233 characters and use no assessment helper calls. The separate Fable 5 artifact has longer professional tasks and output allowances, alongside higher recorded assessment cost.

\paragraph{Separate Fable 5 candidate-cost comparison.}
For GPT-5.5 and Astra, Fable 5's recorded candidate-call cost is 16.73 and 7.12 times Grok's, while its completed-item counts differ by less than 10\% (Table~\ref{tab:cost-same-candidate}).

\begin{table}[ht]
\centering
\small
\begin{tabular}{@{}lrrrr@{}}
\toprule
Candidate & Grok cost & Fable 5 cost & Cost ratio & Completed items \\
\midrule
GPT-5.5 & \$1.5869 & \$26.55 & 16.73$\times$ & 107 / 115 \\
GPT-6 Astra & \$3.5767 & \$25.47 & 7.12$\times$ & 107 / 116 \\
\bottomrule
\end{tabular}
\caption{\textbf{Candidate-matched generation expenditure.} Helper grading is excluded. Completed counts are Grok / Fable 5.}
\label{tab:cost-same-candidate}
\end{table}

\paragraph{Question count versus influence on the score.}
For normalized nonnegative domain weights $w_i$, define the weight-equivalent count
\[
N_w = \frac{1}{\sum_i w_i^2}, \qquad \sum_i w_i=1.
\]
It equals the question count under uniform weights and decreases with concentration. Grok's coding/co-work/reasoning counts are 62/20/25, but $N_w$ is 21.84/6.78/19.07. Fable 5.1's corresponding values are 38/40.32/30 for 38/42/30 questions.

Grok's six highest-weight co-work items carry 93.87\% of that domain score (Figure~\ref{fig:design-concentration}). Low-weight questions still incur execution cost.

\paragraph{Score spread.}
GPT-5.5, Astra, and Sol all score 100\% on Grok co-work, while Fable 5.1 reports 79.22\%, 81.73\%, and 70.52\% for the same candidates. This contrast, alongside Grok's 74.87\% overall PA and 22.59 overall Score, illustrates why task cost, score spread, and target agreement must be assessed separately.

\paragraph{Frozen execution allowances and grading paths.}
The supplied Fable 5.1, Fable 5, and Opus 5 checkpoints match every file hash in their respective freeze manifests. Table~\ref{tab:execution-allowances} summarizes their requested output allowances alongside Grok. The sum is a protocol setting for one full suite, excluding helpers and retries; it is not observed token consumption or a billing bound.

\begin{table}[!htbp]
\centering\small
\begin{tabular}{lrrl}
\toprule
Artifact & Questions & Sum of caps (M tokens) & Per-question caps \\
\midrule
Grok 4.7 & 107 & 0.115 & 768--2,048 \\
Fable 5.1 & 110 & 1.374 & 5,000--16,000 \\
Fable 5 & 118 & 2.261 & 16,384--24,576 \\
Opus 5 & 108 & 1.968 & 8,000--24,000 \\
\bottomrule
\end{tabular}
\caption{\textbf{Declared answer allowances at similar question counts.} A larger allowance permits longer answers and reasoning; it does not require every call to use that many tokens.}
\label{tab:execution-allowances}
\end{table}

Fable 5.1 grades GDPval against upstream rubric criteria using \texttt{helper-grader}; its free-response reasoning uses local matching followed, when needed, by \texttt{helper-simulator}. The corresponding known assessment charges are \$33.9709 and \$0.0070. Fable 5 uses the same division of grader roles, with up to three rubric-judge attempts when a verdict cannot be parsed (\$18.0167 rubric and \$0.0038 equivalence charges). Opus uses deterministic structural/numeric checks for co-work and only the cheap reasoning equivalence path (\$0.0193). Grok has no assessment helper calls. These partitions reconcile to the framework's recorded evaluation totals; outstanding reservations remain separate at \$7.1444, \$1.9747, and \$7.6783 for the three supplied runs.

\paragraph{Observed decisions behind the allowances.}
Fable 5.1's early valid snapshots contain 145--153 items including ArXivMath~\citep{dekoninck2026arxivmath}. Its pilot tool outputs record empty capped ArXivMath responses on both tested candidates. Later snapshots remove that family and reduce the suite to 110 items. A completed development pass then records nine capped GDPval answers out of sixteen, prompting a final increase from 8,000 to 16,000 tokens for that family. Opus's development diagnostic records 36 capped responses out of 102, including 32 empty replies; its final code increases reasoning, coding, and co-work caps to 24,000, 16,000, and 8,000. These diagnostics establish why the researchers changed the settings, without assuming the larger caps necessarily improve every final metric.

Fable 5's trajectory shows that grading is also part of the iteration budget. Its rubric helper initially exhausts its allowance without a parsable verdict; the researcher increases that allowance, adds bounded retry samples, and changes error handling to preserve other item results. Slow candidate calls and late grading failures delay the response matrix needed for weight selection, so the final suite retains its predefined family weights. Monetary spend, feedback latency, and deployment cost are therefore distinct constraints on evaluation research.

%% file: figures/table_execution_cost_components.tex
\begin{table}[!htbp]
\centering\small
\setlength{\tabcolsep}{5pt}
\begin{tabular}{lrrrrr}
\toprule
Artifact & Questions & Candidate & Rubric judge & Equivalence & Total \\
\midrule
Grok 4.7 & 107 & 8.63 & -- & -- & 8.63 \\
Fable 5.1 & 110 & 75.44 & 33.97 & 0.01 & 109.41 \\
Fable 5 & 118 & 100.00 & 18.02 & $<$0.01 & 118.02 \\
Opus 5 & 108 & 105.22 & -- & 0.02 & 105.24 \\
\bottomrule\end{tabular}
\caption{\textbf{Where the execution budget goes.} Recorded assessment expenditure in USD, split by candidate and grader handles. Questions denote declared suite size; costs are incurred run totals, including partial executions, rather than equal-workload estimates. Unresolved reservations are excluded.}
\label{tab:execution-cost-components}
\end{table}

%% file: sections/appendix_ablations.tex
\section{Frozen-Suite Sensitivity and Source-Stratified Baseline}
\label{app:ablation-status}

Smaller question sets cost less to execute, but may change both agreement and measurement variability. To examine this trade-off, the question-count analysis below uses a fixed common panel of twelve candidates and fourteen targets.
\input{figures/table_question_variability}
\input{figures/figure_question_subsampling}

\subsection{Source-stratified Random120 baseline}
\label{app:random120-baseline}
Random120 tests how much agreement can be obtained by sampling existing public tasks without iterative research. We fix source-level quotas to cover multiple task types, then sample within each source independently of target reference scores. The maintainers selected sixteen sources, screened 21,083 eligible items, set source quotas, adapted fixed local graders, and used equal item weights for each raw domain mean. A Python pseudorandom generator with seed 42 sampled only \emph{within} each source. Coding draws eight tasks each from MBPP~\citep{austin2021mbpp}, HumanEval~\citep{chen2021humaneval}, DebugBench~\citep{tian2024debugbench}, LiveCodeBench~\citep{jain2024livecodebench}, and RepoBench~\citep{liu2023repobench}. Co-work draws five from IFEval~\citep{zhou2023ifeval}, six each from BFCL~\citep{patil2025bfcl}, ACPBench~\citep{kokel2026acpbench}, InFoBench~\citep{qin2024infobench}, LIFBench~\citep{wu2025lifbench}, and PlanBench~\citep{valmeekam2023planbench}, and five from BEIR~\citep{thakur2021beir}. Reasoning draws ten each from LogiQA~\citep{liu2020logiqa}, MMLU~\citep{hendrycks2021mmlu}, GPQA~\citep{rein2023gpqa}, and AIME25~\citep{balunovic2025matharena}. The resulting frozen suite has 40 tasks per domain, 120 in total. Twelve candidates completed all three domains, yielding 7--11 matched candidate scores per target across fourteen references. The main-table Random120 row uses these observed scores; the matched comparison recomputes all eight artifacts on this same candidate panel.

\subsection{Matched-panel comparison and score decomposition}
\label{app:matched-panel}
To compare the agent-designed evaluations with Random120 without changing the candidate panel, we recompute all eight complete-domain artifacts on exactly the same twelve candidates and fourteen target-reference intersections as Random120. Every target retains its original score scale and eligibility; no researcher program or output is changed. Table~\ref{tab:matched-baseline} gives all values and the alternative domain-equal aggregate.
\input{figures/table_matched_baseline}

For candidate-panel sensitivity, 10,000 multinomial draws (seed 20260925) resample twelve candidate identities with replacement. The same multiplicities apply jointly to all methods and all references, preserving within-candidate dependence. We recompute population moments and pairwise agreement, weighting candidate pairs by the product of their multiplicities. A common mask retains 9,916 draws with defined metrics for every method and target; the other 84 contain a target with no non-tied reference pair. The figure reports central percentile ranges of paired differences from Random120 on this fixed target set, not research-episode variation or uncertainty from selecting new target families.

For each target, with population standard deviations $s_x,s_y$ and means $\bar x,\bar y$, CCC decomposes exactly as
\[
 \mathrm{CCC}=rC_b, \qquad
 C_b=\frac{2s_xs_y}{s_x^2+s_y^2+(\bar x-\bar y)^2}.
\]
Here $r$ is score correlation and $C_b\in[0,1]$ measures agreement in location and dispersion. With constant measurements we use $r=C_b=0$, preserving CCC zero against a nonconstant reference. Table~\ref{tab:concordance-factors} reports mean per-target factors; their product is not used to approximate macro-CCC. The decomposition describes the frozen score geometry and performs no calibration or label-based fitting.
\input{figures/table_concordance_factors}

Equal weighting of targets and equal weighting of domains answer different aggregation questions because domains contain different numbers of targets. On this matched panel, target-equal weighting gives Grok Score 22.18 and Astra 21.35; domain-equal weighting gives Astra 25.20 and Grok 21.14. The main leaderboard retains its target-equal definition.

\subsection{Question-count protocol}
\label{app:question-subsampling}
Astra, GLM-5.3, and K3 contribute frozen suites of 112, 100, and 107 questions. For each suite, we draw 1,000 nested, domain-stratified subsets of 10, 30, 60, and 90 questions and compare each with its full suite. The same subset is applied to every candidate; retained item weights are renormalized within each domain. Saved completed grades, including wrong and empty answers, are reused without new model calls. Metrics are recomputed per target before target-equal averaging. The percentile bands describe variation across question draws on this fixed panel.

At full size on the twelve-candidate panel, overall (Score, PA) is (21.35, 69.08\%) for Astra, (14.15, 65.01\%) for GLM, and (17.72, 74.17\%) for K3. At 90 questions, 34.4\%, 70.5\%, and 63.3\% of the respective draws fall within one Score point of their full-suite value. These percentages describe within-artifact question subsets, rather than the ordering of different researchers.

\begin{figure}[t]
\centering
\includegraphics[width=\linewidth]{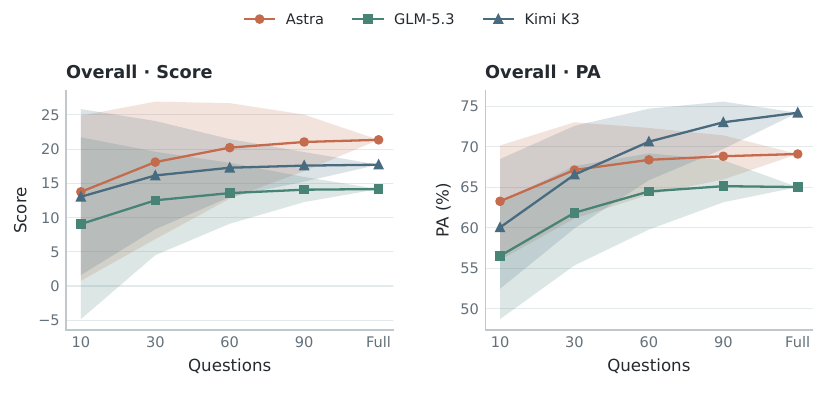}
\caption{Overall Score and PA across nested question subsets. Bands show empirical 2.5--97.5 percentiles over 1,000 draws on the fixed candidate panel.}
\label{fig:question-overall}
\end{figure}

%% file: figures/table_question_variability.tex
\begin{table}[t]
\centering\small
\setlength{\tabcolsep}{5pt}
\begin{tabular}{rrrrrrr}\toprule
Questions & \multicolumn{3}{c}{$\mathrm{SD}(\mathrm{Score})$} & \multicolumn{3}{c}{$\mathrm{SD}(\mathrm{PA})$ (pp)} \\
\cmidrule(lr){2-4}\cmidrule(lr){5-7}
 & Astra & GLM & K3 & Astra & GLM & K3 \\ \midrule
10 & 6.64 & 6.92 & 6.49 & 3.74 & 3.83 & 4.09 \\
30 & 5.48 & 3.97 & 4.06 & 2.97 & 3.20 & 3.24 \\
60 & 3.70 & 2.27 & 2.21 & 2.17 & 2.49 & 2.26 \\
90 & 2.14 & 0.96 & 1.14 & 1.34 & 1.27 & 1.51 \\
\bottomrule\end{tabular}
\caption{\textbf{Metric variability across question subsets.} Overall-score standard deviations over 1,000 draws with stored responses and the twelve-candidate panel held fixed. Score is defined in Section~\ref{sec:quality}; PA variation is in percentage points.}
\label{tab:question-variability}
\end{table}

%% file: figures/figure_question_subsampling.tex
\begin{figure}[t]
\centering
\includegraphics[width=0.92\linewidth]{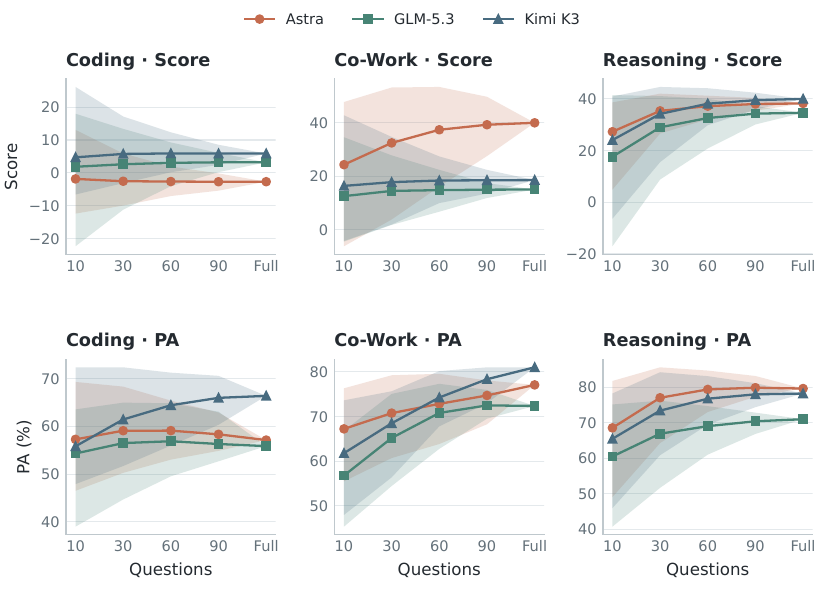}
\caption{\textbf{Question-count sensitivity by domain.} Columns show coding, co-work, and reasoning; rows show Score and PA. Lines and bands give means and empirical 2.5--97.5 percentiles over 1,000 nested domain-stratified subsets on twelve common candidates. Full contains 112/100/107 questions for Astra/GLM/Kimi.}
\label{fig:question-subsampling}
\end{figure}

%% file: figures/table_matched_baseline.tex
\begin{table}[!htbp]
\centering\footnotesize
\setlength{\tabcolsep}{5pt}
\begin{tabular}{lrrrrrr}
\toprule
 & \multicolumn{4}{c}{Target-equal Score} & PA & Domain-equal \\
\cmidrule(lr){2-5}
Artifact & Coding & Co-work & Reasoning & Overall & Overall & Score \\
\midrule
Astra & -2.69 & 40.05 & 38.25 & 21.35 & 69.08 & 25.20 \\
Grok 4.7 & 22.94 & 25.75 & 14.71 & 22.18 & 76.07 & 21.14 \\
Fable 5.1 & 12.79 & 19.89 & 36.87 & 20.48 & 68.46 & 23.18 \\
Opus 5 & -3.83 & 37.18 & 39.78 & 20.16 & 64.81 & 24.37 \\
Kimi K3 & 5.88 & 18.55 & 39.99 & 17.72 & 74.17 & 21.47 \\
Sol & 19.29 & 0.17 & 41.06 & 17.13 & 62.55 & 20.17 \\
GLM-5.3 & 3.20 & 15.02 & 34.62 & 14.15 & 65.01 & 17.61 \\
DeepSeek V4 Pro & 0.59 & 8.92 & 32.72 & 10.45 & 63.51 & 14.08 \\
\midrule
Random120 & 22.92 & 3.75 & 9.03 & 13.10 & 68.29 & 11.90 \\
\bottomrule\end{tabular}
\caption{\textbf{All artifacts and Random120 on the same twelve-candidate panel.} Every method uses the same candidate-reference intersections for each of fourteen targets. Target-equal Score gives each benchmark equal weight; domain-equal Score first averages targets within each domain, then gives each of three domains equal weight. PA is in percent.}
\label{tab:matched-baseline}
\end{table}

%% file: figures/table_concordance_factors.tex
\begin{table}[!htbp]
\centering\footnotesize
\setlength{\tabcolsep}{3.6pt}
\begin{tabular}{lrrrrrrrr}
\toprule
 & \multicolumn{2}{c}{Coding} & \multicolumn{2}{c}{Co-work} & \multicolumn{2}{c}{Reasoning} & \multicolumn{2}{c}{Overall Score} \\
\cmidrule(lr){2-3}\cmidrule(lr){4-5}\cmidrule(lr){6-7}\cmidrule(lr){8-9}
Artifact & $\bar r$ & $\overline{C_b}$ & $\bar r$ & $\overline{C_b}$ & $\bar r$ & $\overline{C_b}$ & Target-eq. & Domain-eq. \\
\midrule
Astra & 0.079 & 0.508 & 0.521 & 0.808 & 0.708 & 0.505 & 21.35 & 25.20 \\
Grok 4.7 & 0.511 & 0.393 & 0.432 & 0.667 & 0.591 & 0.302 & 22.18 & 21.14 \\
Fable 5.1 & 0.321 & 0.498 & 0.328 & 0.614 & 0.633 & 0.527 & 20.48 & 23.18 \\
Opus 5 & 0.103 & 0.510 & 0.451 & 0.817 & 0.695 & 0.524 & 20.16 & 24.37 \\
Kimi K3 & 0.269 & 0.367 & 0.457 & 0.497 & 0.704 & 0.518 & 17.72 & 21.47 \\
Sol & 0.486 & 0.397 & -0.097 & 0.427 & 0.701 & 0.507 & 17.13 & 20.17 \\
GLM-5.3 & 0.191 & 0.405 & 0.514 & 0.275 & 0.585 & 0.517 & 14.15 & 17.61 \\
DeepSeek V4 Pro & 0.124 & 0.157 & 0.384 & 0.185 & 0.561 & 0.505 & 10.45 & 14.08 \\
\midrule
Random120 & 0.529 & 0.480 & 0.091 & 0.740 & 0.409 & 0.335 & 13.10 & 11.90 \\
\bottomrule\end{tabular}
\caption{\textbf{Association, calibration, and aggregation on twelve matched candidates.} Per-target CCC is $rC_b$; $C_b$ penalizes mean and dispersion mismatch. Factor means are reported separately, so $\bar r\,\overline{C_b}$ is not macro-CCC. Overall Score weights either all fourteen targets equally or each domain equally after averaging its targets.}
\label{tab:concordance-factors}
\end{table}

%% file: sections/appendix_research_components.tex
\section{Research Decisions and Native-Component Analysis}
\label{app:research-components}

To determine which parts of a submitted suite contribute to its measurements, we group questions by the task families defined in that submission and remove each group in turn. This component audit reuses saved item responses for twelve candidates with complete scores across Astra, GLM-5.3, and K3, and the same fourteen target references. It uses the fixed score scales and target-equal aggregation in Appendix~\ref{app:matched-comparison}.

For a task group (a \emph{native component}) $C$, we remove its questions, renormalize the remaining frozen weights within that domain, and recompute Score and PA. The reported effect is the full Score minus the Score after deletion. Each effect is compared with 1,000 same-domain random deletions of the same number of questions. These offline comparisons reuse responses and do not rerun changed prompts.

\begin{figure}[p]
\centering
\includegraphics[width=.96\linewidth,height=.86\textheight,keepaspectratio]{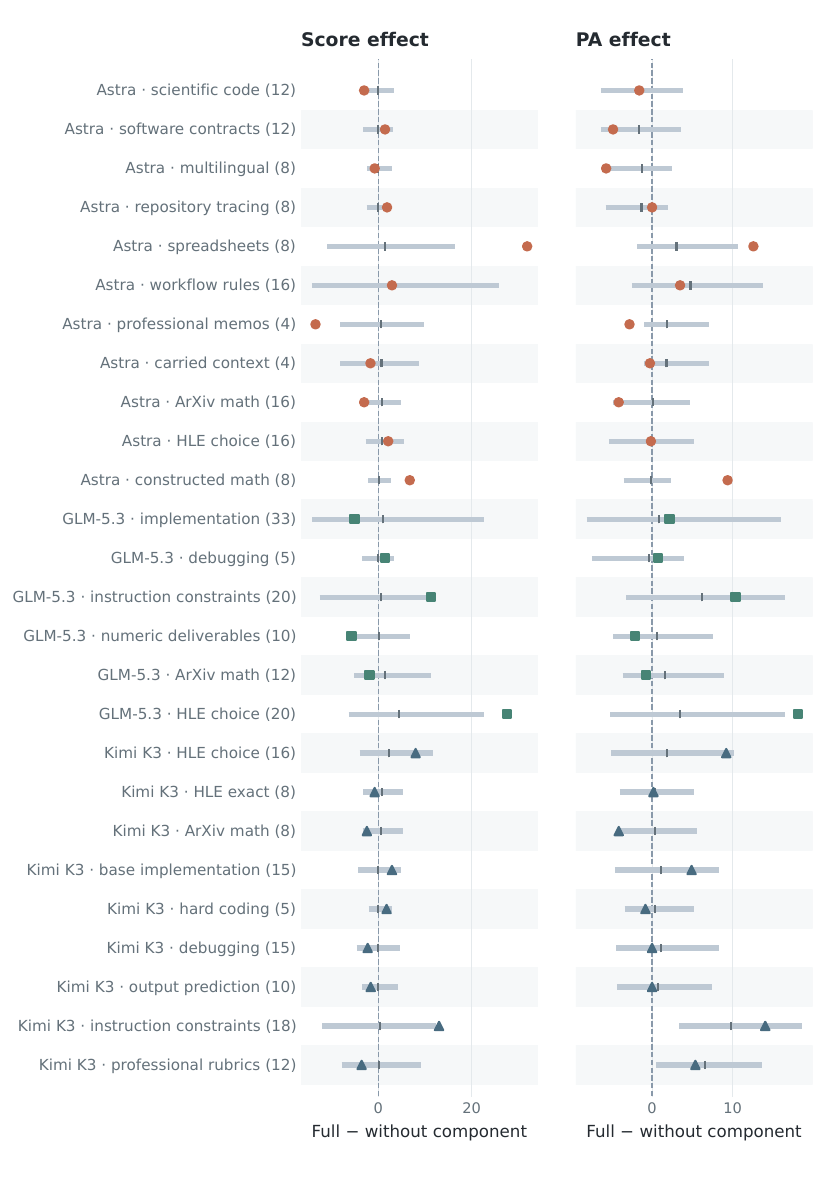}
\caption{All 26 native components. Score effects are in Score points; PA effects are percentage points. Positive values mean removal lowers performance. Gray ranges show central 95\% intervals from matched random removals.}
\label{fig:all-research-components}
\end{figure}

\paragraph{Native grouping.}
Frozen task metadata define eleven Astra, six GLM, and nine K3 disjoint components. Their sizes and observed effects appear in Figure~\ref{fig:all-research-components}.

\begin{figure}[ht]
\centering
\includegraphics[width=.97\linewidth]{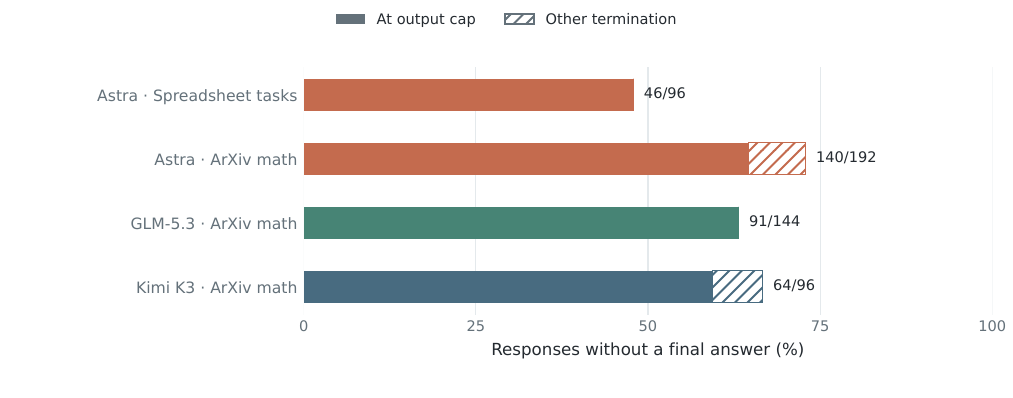}
\caption{Missing final answers in selected components. Solid bars reach the output cap; hatched extensions indicate other termination reasons.}
\label{fig:research-missingness}
\end{figure}

\paragraph{Execution effects.}
Removing Astra's eight spreadsheet tasks reduces co-work Score by 31.94 points and PA by 12.57 percentage points, while 46/96 responses in that component lack final answers at the output cap. Removing the ArXiv-math components~\citep{dekoninck2026arxivmath} raises reasoning Score and PA for all three suites; removing GLM's or K3's HLE choices lowers reasoning PA. Figure~\ref{fig:research-missingness} shows termination counts for selected components.

The complete 26-component analysis appears in Figure~\ref{fig:all-research-components}; all completed wrong and empty grades remain unchanged.

\subsection{Grok: from pilot response patterns to frozen weights}
\label{app:grok-weight-analysis}
Grok first probes hard HLE and coding tasks, encounters truncated responses, and builds a bank of shorter, locally graded questions. The frozen suite contains 62 coding, 20 co-work, and 25 reasoning items. Its saved weight-selection program groups questions by their response patterns on three development candidates. For co-work, the GDPval references are close together. The program allocates most weight to jointly passed items and smaller masses to separating items, searching for the reference ordering and a slope close to one. This search constrains relative differences without matching the reference mean. Six co-work items receive 93.87\% of total weight; the weight-equivalent count $(\sum_iw_i)^2/\sum_iw_i^2$ is 6.78. This quantity describes weight concentration, not the number of statistically independent questions.

The counterfactual uses the same saved grades and candidate--target intersections as the Random120 comparison. It replaces frozen weights with uniform within-domain weights, retaining completed wrong and empty answers and the existing infrastructure exclusions. Recomputed frozen aggregates reproduce all published matched-panel Grok values. Table~\ref{tab:grok-weight-counterfactual} reports every domain, plus a sensitivity check restricted to the ten candidates with all items complete. The comparison changes aggregation only and makes no new model calls.
\input{figures/table_grok_weights}

\begin{figure}[!htb]
\centering
\includegraphics[width=\linewidth]{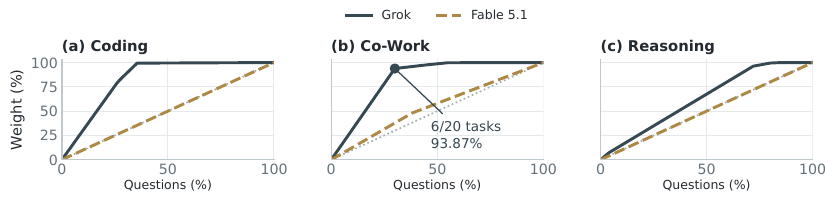}
\caption{\textbf{Weight concentration across domains.} Cumulative scoring mass after sorting questions by decreasing weight, for Grok and Fable 5.1. The diagonal represents uniform weights.}
\label{fig:design-concentration}
\end{figure}

\subsection{DeepSeek V4 Pro: grading convenience and measurement coverage}
DeepSeek's source inspection identifies deterministic grading as a practical way to keep evaluation inexpensive. It constructs Python output-prediction, function-writing, and bug-fixing tasks, plus co-work checks for specified text, numbers, and formats. For example, one co-work question asks for a letter count but supplies the required output string; another asks for time conversion and supplies the converted time. Such checks measure compliance with an explicit response more directly than independent professional problem solving.

The trajectory includes genuine revisions: coding tasks are hardened, a domain-name mismatch in feedback analysis is repaired, and truncated reasoning pilots lead to larger answer allowances and a different question mix. The final reasoning bank includes short-answer HLE questions whose expected answers are easy to parse but whose mathematical content remains demanding. Selecting a short answer simplifies grading, not necessarily reasoning. Final item counts are 38 coding, 51 co-work, and 30 reasoning, with equal within-domain weights.

On the matched panel, DeepSeek's coding and co-work means are high and its concordance calibration factors are low (Table~\ref{tab:concordance-factors}). Its reasoning Score exceeds Random120's, whereas coding accounts for the largest deficit (Table~\ref{tab:matched-baseline}). These observations support the interpretation that simplifying task form can preserve answer checkability while compressing the capability differences the proxy is intended to recover. They do not isolate task content from grading and execution as separate causal interventions.

%% file: figures/table_grok_weights.tex
\begin{table}[!htbp]
\centering\small
\begin{tabular}{llrrrr}
\toprule
 & & \multicolumn{2}{c}{Frozen weights} & \multicolumn{2}{c}{Uniform weights} \\
\cmidrule(lr){3-4}\cmidrule(lr){5-6}
Candidate panel & Domain & Score & PA (\%) & Score & PA (\%) \\
\midrule
Matched twelve & Coding & 22.94 & 73.90 & 36.85 & 73.83 \\
 & Co-work & 25.75 & 77.75 & 50.76 & 78.76 \\
 & Reasoning & 14.71 & 77.62 & 14.53 & 75.76 \\
 & Overall & 22.18 & 76.07 & 37.03 & 76.01 \\
\midrule
Complete ten & Coding & 18.76 & 70.88 & 34.61 & 75.14 \\
 & Co-work & 31.65 & 82.13 & 51.30 & 82.14 \\
 & Reasoning & 11.07 & 83.36 & 12.67 & 83.81 \\
 & Overall & 21.71 & 77.57 & 35.87 & 79.50 \\
\bottomrule\end{tabular}
\caption{\textbf{Grok aggregation counterfactual.} The matched panel uses the same twelve candidates as the Random120 comparison. The complete-item panel retains the ten candidates with all 107 grades available. Targets and within-panel item exclusions remain fixed between variants. These are offline counterfactuals; the primary leaderboard retains the submitted weights.}
\label{tab:grok-weight-counterfactual}
\end{table}

%% file: sections/appendix_prompts.tex
\section{Research Instructions and Inputs}
\label{app:prompts}

Researchers are asked to design one executable evaluation with at least 100 distinct scored questions spanning coding, co-work, and reasoning. They may select, adapt, or synthesize tasks and choose graders, weights, grouping, and output allowances. During development, researchers compare directly measured candidate-domain scores with visible target-reference scores; final assessment reports Score and PA for frozen programs.

The supplied workspace contains the task contract, development candidate APIs and available reference scores, original benchmark materials, available trajectories, and a Python evaluation SDK. The SDK submits a suite, records item-level responses and costs, and returns cross-model feedback on completed development tests. Researchers decide what to test, revise, and freeze within the three-hour research period, \$100 researcher wallet, and \$100 shared development/helper wallet. Frozen assessment has a separate \$120 wallet and \$30 per-candidate cap.

The final program supplies executable questions, grading rules, and three fixed domain aggregations. Completed wrong or empty answers retain their grades; infrastructure failures remain incomplete. After freezing, the executor runs the candidate panel without sending assessment results back to the researcher.

%% file: sections/appendix_research_case.tex
\section{Case Study: How Researchers Revise Evaluations}
\label{app:research-case}

The aggregate analyses show which design choices affect measurement, but do not explain how agents arrive at them. We now follow the saved research trajectories from source inspection through pilot testing and revision. Astra provides a concrete example of this loop; GLM and K3 illustrate different responses to development feedback.

\paragraph{Astra turns source inspection into a coverage hypothesis.}
A compact coding evaluation can miss repository-level behavior if it concentrates on isolated implementations. After inspecting upstream repository-investigation tasks, Astra identifies a gap in its implementation-heavy coding bank and constructs eight multi-file repository traces. In one, candidates predict cached outputs and recomputation counts after in-place mutation versus replacing a dependency object. Running the supplied fixture generates the expected outputs for grading (Figure~\ref{fig:astra-iteration}a).

\paragraph{Pilot feedback can expose defects in the evaluation itself.}
A low pilot score may expose an unclear question or an overly restrictive grader, as well as a candidate error. Astra examines these possibilities in a performance-plan memo that scores 11/12: its response reports that 3/13 orders finish on time, but omits the mean close time required by the rubric. Astra finds that the question never explicitly requested that statistic and adds it to the brief. A later audit also relaxes a cloud-migration rubric that prescribed one architecture despite an open-ended brief, retaining the stated security and delivery requirements (Figure~\ref{fig:astra-iteration}b,c). These revisions align prompts and graders before freezing; the saved pilot score precedes the repairs.
\input{figures/figure_astra_iteration}

\paragraph{Researchers make different revisions from similar feedback.}
GLM corrects invalid examples and numerical answers; K3 responds to near-ceiling pilots with five hard coding tasks. Astra also changes answer allowances and grouped execution after truncated responses. These examples show three uses of pilot feedback: correcting task validity, extending coverage, and changing execution settings. The component analysis in Section~\ref{sec:design-analysis} tests whether the resulting task groups improve agreement on the frozen candidate panel.

%% file: figures/figure_astra_iteration.tex
\begin{figure}[!htb]
\centering
\includegraphics[width=\linewidth]{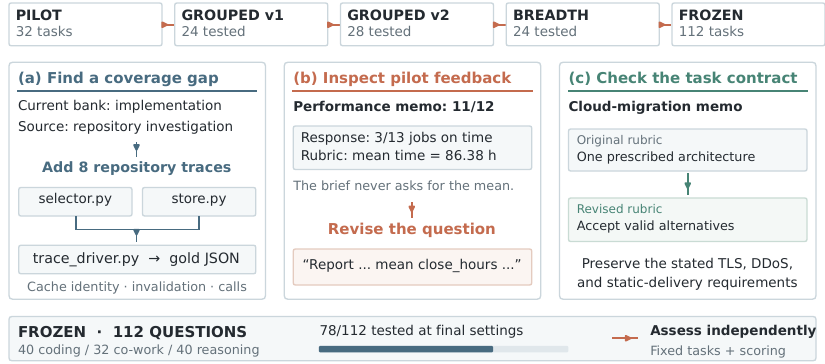}
\caption{\textbf{Astra's research loop.} Source inspection motivates repository-tracing tasks; an 11/12 pilot exposes an implicit requirement, prompting a clearer question; a later validity audit broadens an overly restrictive rubric. The top strip shows saved versions (batch counts are not cumulative). The 112-question submission is then frozen for independent assessment.}
\label{fig:astra-iteration}
\end{figure}